\documentclass[letterpaper, 10 pt, conference]{ieeeconf} 
\IEEEoverridecommandlockouts
\usepackage{siunitx}
\usepackage{physics}
\usepackage[T1]{fontenc}
\AtBeginDocument{\RenewCommandCopy\qty\SI}
\usepackage{amsmath,amsfonts}
\usepackage{array}
\usepackage{cite}
\usepackage[caption=false,font=scriptsize]{subfig}
\usepackage{textcomp}
\usepackage{stfloats}
\usepackage{float}
\usepackage{url}
\usepackage{verbatim}
\usepackage{graphicx}
\usepackage{amssymb}
\usepackage{pifont}
\usepackage{algorithm}
\usepackage{algpseudocode}
\usepackage[disable]{todonotes} 
\usepackage[hidelinks]{hyperref}
\usepackage[none]{hyphenat}
\usepackage[capitalize]{cleveref}
\usepackage{comment}
\usepackage{multirow}
\usepackage{booktabs}
\usepackage{rotating}
\usepackage{makecell}
\usepackage{tabularx}
\usepackage{xcolor}
\usepackage{blindtext}
\usepackage{algorithm}
\usepackage{algpseudocode}
\usepackage{pgf}
\usepackage[utf8]{inputenc}
\usepackage[T1]{fontenc}
\usepackage[inkscapelatex=false, inkscapepath=./build/svg-inkscape]{svg}

\def\BibTeX{{\rm B\kern-.05em{\sc i\kern-.025em b}\kern-.08em
    T\kern-.1667em\lower.7ex\hbox{E}\kern-.125emX}}
    
\begin{document}

\title{\LARGE \bf
A Quasi-Steady-State Black Box Simulation Approach for the Generation of g-g-g-v Diagrams}

\author{
Frederik Werner, Simon Sagmeister, Mattia Piccinini, Johannes Betz%
\thanks{F. Werner and S. Sagmeister are with the Institute of Automotive Technology, TUM School of Engineering and Design, Technical University of Munich, 85748 Garching, Germany; Munich Institute of Robotics and Machine Intelligence (MIRMI), corresponding author: frederik.werner@tum.de.}%
\and
\thanks{M. Piccinini and J. Betz are with the Professorship of Autonomous Vehicle Systems, TUM School of Engineering and Design, Technical University of Munich, 85748 Garching, Germany; Munich Institute of Robotics and Machine Intelligence (MIRMI)}
}

\onecolumn

\begin{center}
    \textcopyright \ 2025 IEEE. Personal use of this material is permitted. Permission from IEEE must be obtained for all other uses, including reprinting/republishing this material for advertising or promotional purposes, collecting new collected works for resale or redistribution to servers or lists, or reuse of any copyrighted component of this work in other works.
\end{center}

\twocolumn

\maketitle


\begin{abstract}

The classical g-g diagram, representing the achievable acceleration space for a vehicle, is commonly used as a constraint in trajectory planning and control due to its computational simplicity. To address non-planar road geometries, this concept can be extended to incorporate g-g constraints as a function of vehicle speed and vertical acceleration, commonly referred to as g-g-g-v diagrams. However, the estimation of g-g-g-v diagrams is an open problem. Existing simulation-based approaches struggle to isolate non-transient, open-loop stable states across all combinations of speed and acceleration, while optimization-based methods often require simplified vehicle equations and have potential convergence issues.
In this paper, we present a novel, open-source, quasi-steady-state black box simulation approach that applies a virtual inertial force in the longitudinal direction. The method emulates the load conditions associated with a specified longitudinal acceleration while maintaining constant vehicle speed, enabling open-loop steering ramps in a purely QSS manner. Appropriate regulation of the ramp steer rate inherently mitigates transient vehicle dynamics when determining the maximum feasible lateral acceleration. Moreover, treating the vehicle model as a black box eliminates model mismatch issues, allowing the use of high-fidelity or proprietary vehicle dynamics models typically unsuited for optimization approaches.
An open-source version of the proposed method is available at: https://github.com/TUM-AVS/GGGVDiagrams

\end{abstract}

\section{INTRODUCTION}
The g-g diagram visualizes the maximum longitudinal and lateral accelerations that a vehicle can achieve. This diagram has historically been widely used in race car driver analysis, and is an essential tool in vehicle dynamics for driver development and setup optimization. Minimizing lap times necessitates maximizing acceleration, which can be compared using the area enclosed by the g-g diagram \cite{Rice.1973, Volkl.2013, Kritayakirana.2010, Brayshaw.2005, Tremlett.2014}.

In the context of autonomous motorsport, g-g diagrams are often employed as constraints for offline trajectory optimization\cite{Novi.2020, Veneri.2020, Massaro.2021, Lovato.2022, Fu.2018}, real-time trajectory planning \cite{Pagot.2020, Rowold., Montani.2021b, Piccinini.2023b, Piccinini.2023, Piccinini.2024b, Piccinini_ggv.2024, Piccinini_primitives_2024} and control \cite{Wischnewski.2023, Novi.2020, Subosits.2019, Fu.2018, Funke2017} algorithms to account for the vehicle's dynamic limitations. Computationally simple point-mass models are an attractive solution, particularly in real-time-critical applications. While linear and nonlinear single-track models with simplified tire models are common alternatives, they are more demanding both computationally and in terms of convergence.
This paper deals with the problem of generating g-g diagrams and their extensions considering the vehicle speed and vertical acceleration.

\begin{figure}
\centering
\includegraphics[width=1\columnwidth]{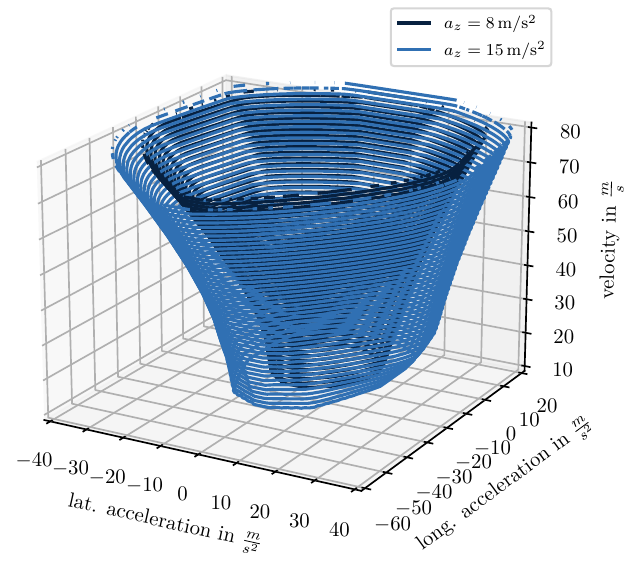}
\caption{Representation of the four-dimensional g-g-g-v diagram visualized at a vertical acceleration $a_z$ of $8$ and $15$ \(\frac{\mathrm{m}}{\mathrm{s}^2}\). This g-g-g-v diagram was generated with the proposed method, using the validated vehicle model of Section \ref{sec:two_track_model_aero}.}
\label{fig:example_gggv}
\end{figure}

\subsection{Extension of the g-g Diagram Considering Vehicle Speed and Vertical Acceleration}

Both the vehicle speed and the vertical acceleration affect the shape of the g-g diagrams. The vehicle speed impacts the diagram mainly due to aerodynamic influences; higher speeds increase aerodynamic forces such as drag and downforce, affecting the acceleration capabilities. The vertical acceleration varies in non-planar, 3D track geometries depending on the track's elevation changes and the vehicle's velocity vector. These effects are evident when driving over crests, through dips, or on segments with road inclinations.

By utilizing 3D track formulations, these effects can be accounted for, providing a more accurate representation of the vehicle's dynamic limitations. Extending the g-g diagram to include vehicle speed $v$ and vertical acceleration $a_z$ leads to \textit{g-g-g-v} diagrams, an example of which is depicted in Fig. \ref{fig:example_gggv}. g-g-g-v diagrams are especially valuable in complex driving scenarios involving non-planar driving surfaces and aerodynamic effects, where considering these additional factors leads to improved accuracy in simulations and better-informed decisions in trajectory planning and control strategies \cite{Rowold., Lovato.2022}.

We will first review the existing methods for generating g-g-g-v diagrams, followed by a presentation of our contributions.
\subsection{Related Work}
Current approaches for generating g-g and g-g-g-v diagrams predominantly rely on data from simulated experiments \cite{Fu.2018, Montani.2021b, Piccinini.2023b} or on direct quasi-steady-state (QSS) optimization techniques \cite{Brayshaw.2005, Tremlett.2014, Massaro.2023, Rowold., Veneri.2020, Massaro.2021, Lovato.2022}. 

However, the simulated-experiments approach typically struggles to ensure full coverage of the whole g-g surface, or to provide guarantees regarding non-transient reachability and open-loop stability \cite{Piccinini.2024b}.

On the other hand, all optimization-based approaches for generating g-g diagrams share the common disadvantage of requiring an adapted formulation of the vehicle system equations. This formulation must be suitable for optimization with respect to differentiability, numerical stability, and complexity. Consequently, a simplified version of an existing vehicle model often needs to be used. In larger software projects, there typically exists a continuously developed and validated high-fidelity vehicle model that serves as a closed-loop reference simulation model. Adopting a lower-complexity version of the reference model leads to increased model maintenance effort, and model mismatch problems cannot be ruled out. Another downside of optimization-based techniques is the chance of finding poor local minima, which may not represent the maximum vehicle performance.

Hence, the development of a method that can directly use existing complex vehicle models without necessitating simplification would reduce maintenance overhead and minimize the risk of discrepancies between models, ultimately enhancing the efficiency and accuracy of the g-g-g-v diagram generation. This way, complex models not suitable for optimization can be used, leveraging all the available model information.

\subsection{Contributions}
The g-g-g-v diagram can be derived either by fitting from simulated experiments, or through optimization techniques.
To overcome the limitations of the existing techniques, we introduce a novel simulation-based approach to generate QSS g-g-g-v diagrams. In summary, this work has three main contributions:

\begin{enumerate} 
\item We provide an open-source, parallelized black box simulation toolchain to generate QSS g-g-g-v diagrams~\cite{Werner.2025}. Our approach enables the use of existing high-fidelity vehicle models without the need for differentiability, avoiding model mismatch problems. 
\item We introduce a novel virtual inertial force concept that opposes and balances the effect of the required longitudinal tire forces for a specified longitudinal acceleration. This approach decouples vehicle speed variation from longitudinal acceleration inertial effects, enabling quasi-steady steering ramps at a constant velocity even under nonzero longitudinal acceleration.
\item This work offers a solution to exclude open-loop unstable driving situations from the g-g-g-v diagram, ensuring open-loop vehicle stability within the derived diagram area. 
\end{enumerate}

\section{METHOD}\label{sec:method}
This section describes our QSS framework to generate g-g-g-v diagrams with black box vehicle models. We begin with a review of the Milliken Moment Method, whose principle is used in our g-g-g-v generation algorithm. We then proceed with a conceptual overview of the proposed technique, outlining how the application of external inertial forces enables constant-speed steering ramps under nonzero longitudinal acceleration. Next, we detail the integration of these forces into a black box vehicle model, highlighting how open-loop stability is preserved and transient effects are minimized. Finally, we provide a summary of the core algorithmic steps, and we describe the integration of our vehicle model.

\begin{figure}
\centering
\includegraphics[width=1\columnwidth]{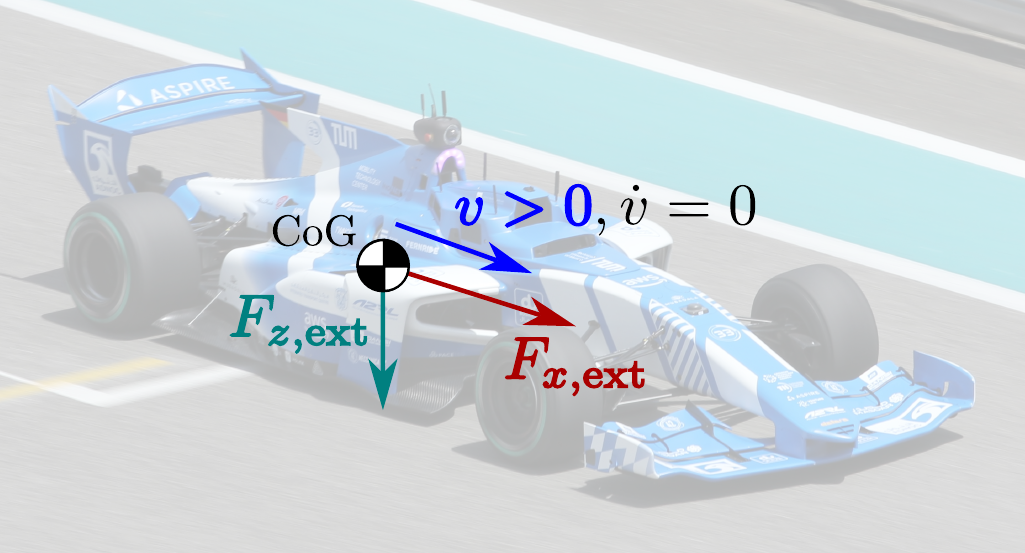}
\caption{Virtual longitudinal and vertical external forces $(F_{x,\mathrm{ext}},F_{z,\mathrm{ext}})$ used in our simulation-based generation of the g-g-g-v diagram. We perform ramp steer maneuvers, using a longitudinal controller to counterbalance $F_{x,\mathrm{ext}}$ and keep a constant vehicle speed $v$, thus ensuring QSS conditions. In Section \ref{sec:two_track_model_aero}, we apply our algorithm to a high-fidelity model of the autonomous race car depicted in the image.}
\label{fig:external_forces}
\end{figure}

\subsection{Conceptual Overview of the Proposed Method}
The developed approach relies on the Milliken Moment Method, which is outlined in the next section, and treats the vehicle model as a \emph{black box}. This means model integration is purely forward and no access to internal model states or differentiability is required. 

Instead, external longitudinal and vertical forces are imposed at the vehicle’s center of gravity (CoG) as depicted in Fig. \ref{fig:external_forces}. These forces emulate inertial effects, and are counterbalanced by a wheel torque controller to keep the vehicle speed constant. Thus, we preserve QSS conditions, but with identical tire loads and tire slips that would arise under a specified longitudinal acceleration \(a_x\).
This approach enables open-loop steering ramps to be conducted at a desired speed \(v\), yet with variable longitudinal and vertical external forces,
to replicate the tire load distribution and slip behavior consistent with nonzero \(a_x\).

Powertrain torque limits are omitted to focus on \emph{tire-force limitations} only, disentangling powertrain constraints from tire constraints. This separation simplifies the analysis and allows for an independent assessment of the tire-force envelope.

\begin{figure}
\centering
\includegraphics[width=1\columnwidth]{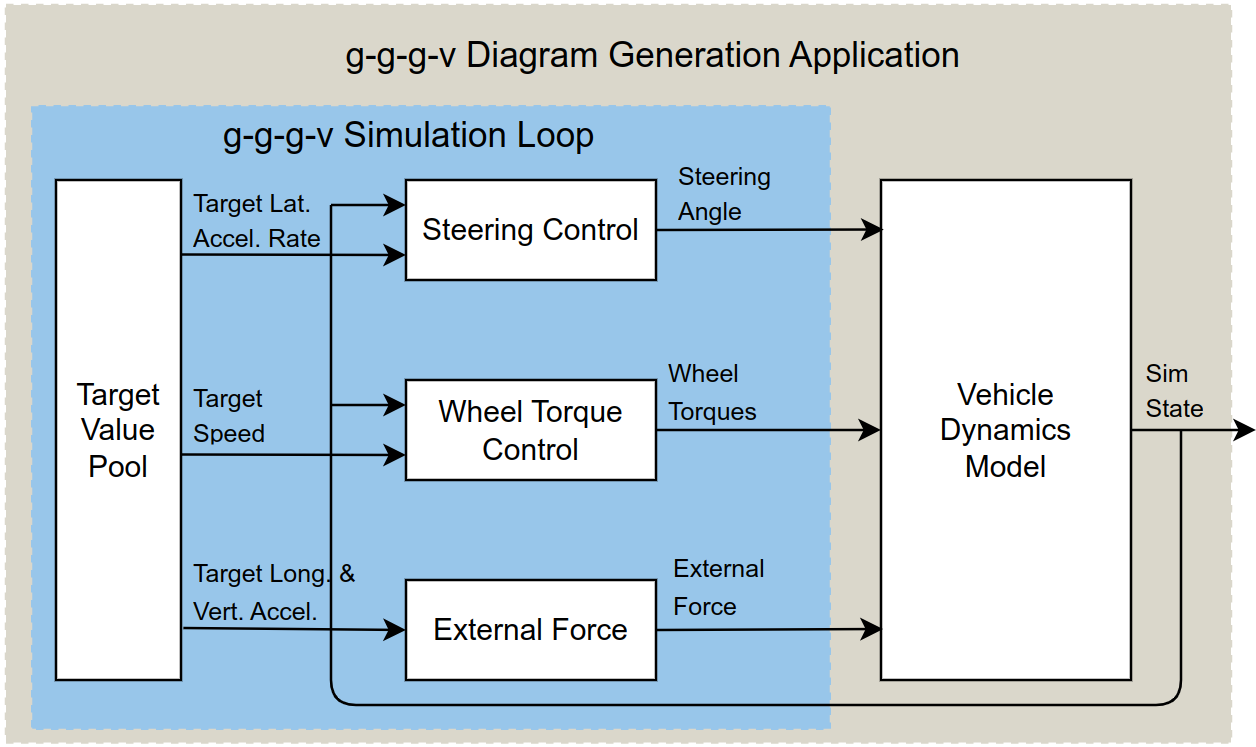}
\caption{Block diagram of the QSS black box simulation approach.}
\label{fig:block_diagram}
\end{figure}

\subsection{Principle of the Milliken Moment Method}
Originally rooted in Perkins' wind tunnel experiments on aircrafts, the Milliken Moment Method (MMM) \cite{Milliken.1995} provides a systematic way to characterize the forces and moments acting on a vehicle under various operating conditions. Although the MMM is commonly employed to generate yaw-moment diagrams used to analyze understeer/oversteer behavior, its fundamental idea can be seen as closely related to generating performance envelopes such as g-g or g-g-g-v diagrams. In particular, the shared principle lies in conducting QSS analyses where one or more degrees of freedom are constrained, thus allowing the vehicle’s reaction forces and moments to be measured for different combinations of steering angle, acceleration, and speed.

While the MMM was originally described as a cable-based physical test rig described in \cite[Chapter 8]{Milliken.1995}—constraining the vehicle in specific directions to measure forces and moments—it can be replicated more easily in simulation by imposing external forces and moments on the virtual vehicle’s center of gravity. The MMM approach naturally captures realistic tire force development, since the vehicle model is still free to undergo the kinematic movements like roll, pitch, vertical wheel travel that dictate vertical load transfer and tire slip. In addition, balancing external forces in the longitudinal direction enables the emulation of inertial forces and acceleration without modifying the vehicle’s velocity, effectively simulating traction or braking forces needed for a given acceleration state without changing the nominal driving speed.

\subsection{QSS Black Box Simulation through External Inertial Forces}
Fig.~\ref{fig:block_diagram} provides an overview of our simulation framework. For each combination of vehicle speed $v$ and vertical acceleration $a_z$, a range of longitudinal accelerations \(\{a_{x,i}\}\) (target value pool)
is tested and a ramp steer maneuver is performed. As shown in Fig. \ref{fig:external_forces}, the accelerations $(a_x, a_z)$ are superimposed by means of virtual external forces $(F_{x,\mathrm{ext}},F_{z,\mathrm{ext}})$:
\begin{equation}
F_{x,\mathrm{ext}} = - m \, a_x, 
\quad
F_{z,\mathrm{ext}} = m\bigl(a_z - g\bigr),
\label{eq:ext_forces}
\end{equation}
where \(m\) is the vehicle mass, and \(g\) is the gravitational acceleration. Although \(F_{x,\mathrm{ext}}\) implies a nonzero longitudinal acceleration \(\dot{v} = \dv{v}{t} = a_x\) under normal circumstances, we enforce \(\dot{v}=0\) through a PID wheel-torque controller that keeps $v$ constant. This controller counterbalances any net longitudinal force (including aerodynamic drag and rolling resistance) to hold the speed at \(v\), allowing QSS ramp steer maneuvers to be performed around each snapshot \(\bigl(v,a_x,a_z\bigr)\) without introducing significant transient effects. Consequently, the net longitudinal acceleration of the vehicle is zero, but, from the perspective of load transfer and tire slip, the vehicle experiences the same tire loads as it would at an acceleration of \(\dot{v} = a_x\). Note the minus sign in the expression of \(F_{x,\mathrm{ext}}\) in \eqref{eq:ext_forces}: when we want to emulate the effects of a positive longitudinal acceleration $a_x$, in our setting we need to apply a negative external force to the vehicle's CoG. In the case of positive emulated longitudinal acceleration, the drive torque controller actuates the driven axle only, in case of negative emulated accelerations the drive torque is distributed according to the brake balance onto all four wheels.

Since our approach treats the vehicle model as a black box, the method remains compatible with complex or proprietary models that may be unsuitable for optimization-based approaches.

Due to the independence of each simulation run, the method allows for parallel execution.
The total runtime is inversely proportional to the number of available processor cores.

\subsection{Quasi-Steady-State Ramp Steer Maneuver}
To ensure quasi-steady-state operation of the vehicle model during the ramp steer, a two-step steering method is employed:
\begin{enumerate}
    \item \emph{Lateral Acceleration Gain Test:}  
    A small steering step \(\Delta\delta\) is applied to measure the resulting \(\Delta a_y\). From
    \begin{equation}
        \kappa = \frac{\Delta a_y}{\Delta\delta},
        \label{eq:lat_accel_gain}
    \end{equation}
    we compute the \emph{lateral acceleration gain $\kappa$}. Section \ref{sec:lat_accel_gain} will show that this gain strongly depends on $a_x$ and $v$. A target lateral acceleration ramp rate of about \(1\,\mathrm{m}/\mathrm{s}^3\) was empirically found in a sensitivity sweep to approximate QSS behavior with less than \(0.1\%\) error compared to a very slow, near-infinite ramp and offers a good trade off between accuracy and runtime. Hence, the required steering rate \(\dot{\delta}\) becomes:
    \begin{equation}
      \dot{\delta} = \frac{1\,\mathrm{m}/\mathrm{s}^3}{\kappa}.
    \end{equation}

    \item \emph{Ramp Steer Execution:}  
    The steering angle is continuously increased at \(\dot{\delta}\). The vehicle model develops slip angles and tire forces.
\end{enumerate}

\subsection{Detecting the Open-Loop-Stable Maximum Lateral Acceleration} \label{subsec:detection}
Each ramp steer simulation will either reach front axle or rear axle saturation. Front axle saturation results in limit understeer which is trivial to find, however for rear axle saturation a method to detect the change from a quasi-steady to a transient vehicle state must be introduced.

\textbf{Front-Axle Saturation (Understeer).}  
A local maximum in \(a_y(t)\) indicates the peak lateral acceleration before understeer reduces cornering capability. This peak is taken as \(a_{y,\text{max}}\).

\textbf{Rear-Axle Saturation (Oversteer).}  
Oversteer is detected by comparing \(a_y\) with \(v\,\dot{\psi}\) in steady-state cornering, with $\dot{\psi}$ being the yaw rate. When
\begin{equation}
  \bigl|a_y - v\,\dot{\psi}\bigr| \;>\; \varepsilon_{\text{th}},
\end{equation}
\label{eq:oversteer_detection}
the vehicle has begun to rotate faster than the lateral acceleration would suggest in steady-state, and the last stable sample is stored as the maximum open-loop lateral acceleration \(a_{y,\text{max}}\).

\textbf{Side slip Angle Correction.}  
After identifying the maximum lateral acceleration \(a_{y,\text{max}}\) in the vehicle frame, we transform it to the velocity vector reference frame:
\begin{equation}
a_{y,\text{corr}} = a_{y,\text{max}}\cos\bigl(\beta\bigr) - a_{x} \sin(\beta),
\end{equation}
where \(\beta\) is the side slip angle. This ensures that the final reported lateral acceleration is orthogonal to the velocity vector, which makes the resulting g-g-g-v diagram more applicable as a constraint for point mass models like the ones of \cite{Rowold.,Novi.2020,Veneri.2020}, which fundamentally lack the slip angle as a system state.

\subsection{Algorithmic Overview}
Algorithm~\ref{alg:accel_limits} summarizes the method described above. The simulation runs are executed in parallel and the resulting g-g-g-v points are stored in a 4D matrix.

\begin{algorithm}
\caption{QSS Ramp Steer (Parallel Execution)}
\label{alg:accel_limits}
\begin{algorithmic}[1]
    \State \textbf{Input:} \textit{speed} \((v)\), \textit{vertical accel.} \((a_z)\), \textit{longitudinal accel.} \((a_x)\)
    \ForAll{\( (v, a_z, a_x)\)}
        \State Initialize vehicle speed to \(v\)
        \State Apply \(F_{z,\mathrm{ext}} = m\bigl(a_z - g\bigr)\)
        \State Apply \(F_{x,\mathrm{ext}} = - m\,a_x\)
        \State Use PID control to hold \(v\) (wait for steady state)
        \State Apply small steering step \(\Delta\delta\)
        \State Measure \(\Delta a_y\), compute lateral gain \(\kappa\)
        \State Calculate \(\dot{\delta} = \tfrac{1\,\mathrm{m}/\mathrm{s}^3}{\kappa}\)
        \State Ramp steering at \(\dot{\delta}\)
        \If{Ramp steer done}
            \State $a_{y,\text{max}} = \max(a_{y}(t)) $ 
        \ElsIf{open-loop instability (\(\lvert a_y - v\dot{\psi}\rvert > \varepsilon_{\text{th}}\))}
        \State $a_{y,\text{max}} = a_{y}(t \,\, \text{where} \,\, \lvert a_y - v\dot{\psi} \rvert > \varepsilon_{\text{th}})$
        \EndIf
        \State $a_{y,\text{corr}} =  a_{y,\text{max}}\cos\bigl(\beta\bigr) - a_{x} \sin(\beta)$
    \EndFor
    \State \textbf{Output:} Map of $a_{y,\text{corr}}$ for each $(v, a_z, a_x)$
\end{algorithmic}
\end{algorithm}

\subsection{Vehicle Dynamics Simulation}

We aim to derive the acceleration limits for real test vehicles. However, we do this with a simulation-based approach. 
We chose the high-fidelity, validated, open-source vehicle model "Open Car Dynamics" \cite{sagmeister2024}. This model can accurately capture the dynamics of the real full-scale race car shown in Fig. \ref{fig:external_forces}, even close to its handling limits. 

\begin{figure}[!tb]
\centering
\includegraphics[width=0.9\columnwidth]{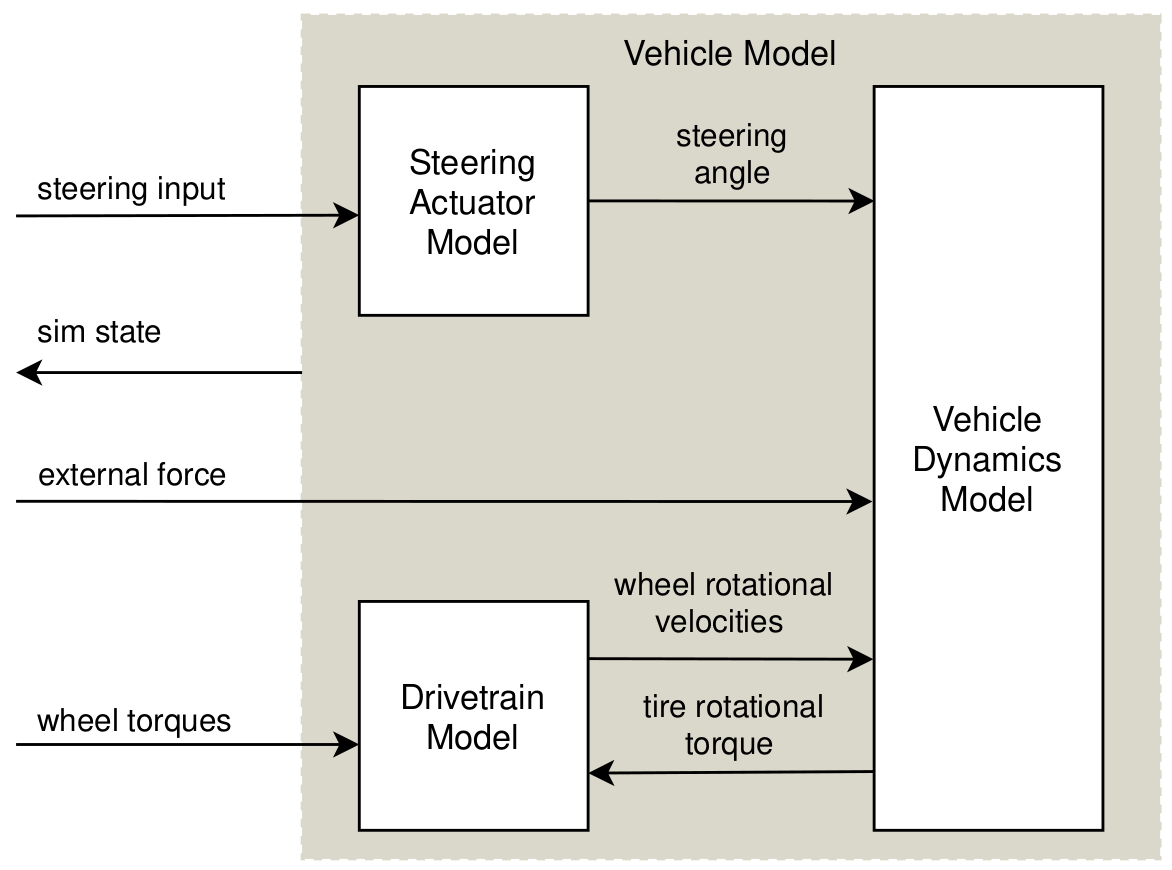}
\caption{Structure of the "Open Car Dynamics" vehicle model according to \cite{sagmeister2024}. Each vehicle is composed of a vehicle dynamics, a steering actuator, and a drivetrain model.}
\label{fig:model-structure}
\end{figure}

The race car used to collect the data for real-world validation is controlled using steering angle, throttle, and brake pressure as input commands. However, as mentioned previously, our approach requires a drive torque for each wheel as input. Here, we leverage the model's modular structure shown in Fig. \ref{fig:model-structure}. Since the drivetrain is a separate module, it can easily be replaced with a different implementation. By creating a drivetrain model that takes 4 individual wheel torques as inputs, we can create a vehicle that fits our input constraints, while keeping the already validated vehicle dynamics model untouched. 

\section{RESULTS} \label{sec:results}
This section presents the key outcomes of our QSS black box approach. We first show how the ramp steer strategy identifies maximum open-loop stable accelerations, distinguishing understeer and oversteer conditions. We then validate the method’s accuracy against a simple force-constrained vehicle model, and finally demonstrate its applicability to an extended high-fidelity model with variable aerodynamics.

\subsection{Ramp Steer and Lateral Acceleration Gain} \label{sec:lat_accel_gain}
\begin{figure}
\centering
\includegraphics[width=0.95\columnwidth]{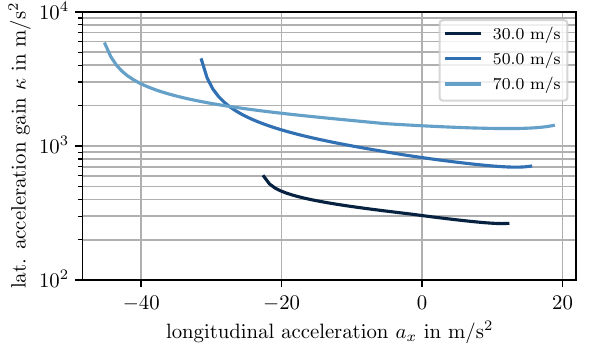}
\caption{Lateral acceleration gain $\kappa$ (defined in \eqref{eq:lat_accel_gain}) for different speeds $v$ and longitudinal acceleration $a_x$. Adapting $\dot{\delta}$ based on $\kappa$ preserves QSS conditions, ensuring quasi-steady vehicle behavior throughout the ramp steer maneuver and reliable oversteer detection.}
\label{fig:lat_acc_gain}
\end{figure}

The following section examines the distinction between understeer and oversteer scenarios, and the importance of adaptively regulating the ramp steer rate \(\dot{\delta}\) for each simulation run with different \(\bigl(v,a_x,a_z\bigr)\). To illustrate these points, we analyze Fig. \ref{fig:steer_ramp_subplots} and Fig. \ref{fig:lat_acc_gain}. The first figure contains three plots:

\begin{figure}
\centering
\includegraphics[width=0.95\columnwidth]{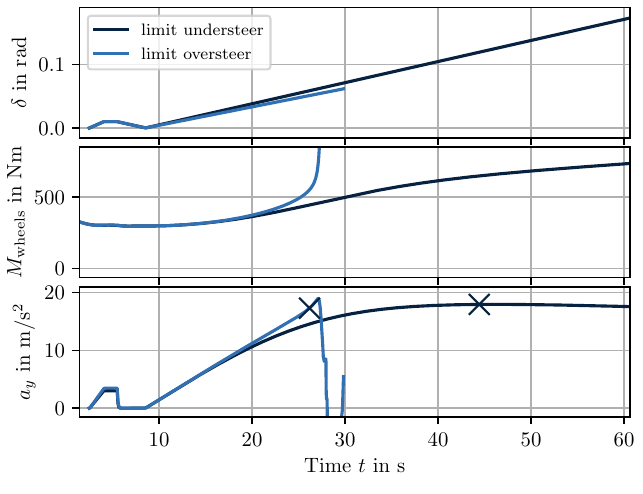}
\caption{Top: Step steer test followed by ramp steer. Middle: Corresponding wheel torque evolution. Bottom: Lateral acceleration for understeer and oversteer cases for a rear-wheel-driven race car at a velocity of $30\,\mathrm{m/s}$ and target $a_x$ of $0\,\mathrm{m/s^2}$.
The detected open-loop stable maximum $a_{y}$ is marked with a black cross as described in Section~\ref{subsec:detection}.}
\label{fig:steer_ramp_subplots}
\end{figure}

\begin{itemize}
    \item \textbf{Steering Input and Torque} (first two plots of Fig. \ref{fig:steer_ramp_subplots}):  
    A small step steer is initially applied to determine the lateral acceleration gain (see Section~\ref{sec:method}), followed by the full ramp steer. The wheel torque naturally increases as the tires experience higher lateral slip and rolling resistance.
    \item \textbf{Lateral Acceleration Response} (bottom plot of Fig. \ref{fig:steer_ramp_subplots}):  
    The figure shows two runs: one exhibiting limit understeer and the other, limit oversteer. In each case, our method identifies the maximum open-loop stable lateral acceleration as described in Section~\ref{sec:method}. The understeering vehicle exhibits a clear peak in $a_{y}$, while the oversteering vehicle transitions into a open-loop unstable region, detected by \ref{eq:oversteer_detection}
\end{itemize}

Fig.~\ref{fig:lat_acc_gain} shows the measured lateral acceleration gain $\kappa$ (defined in \eqref{eq:lat_accel_gain}) for different velocities $v$ and longitudinal accelerations $a_x$. The plot shows that $\kappa$ varies significantly as a function of $v$ and $a_x$. Thus, adapting the ramp steer rate $\dot{\delta}$ based on $\kappa$ is crucial to ensure quasi-steady behavior and reliable oversteer detection.
By carrying out a small step steer test before each simulation, the appropriate ramp steer rate $\dot{\delta}$ is determined adaptively, remaining sufficiently slow to preserve quasi-steady behavior. Consequently, the oversteer detection method becomes more robust, since any abrupt yaw motion stands out clearly when the rest of the vehicle state is nearly stationary.

\subsection{Validation against a Force-constrained Model with an Analytical Solution}
For validation, our goal is to prove the accuracy of our method rather than the accuracy of the utilized vehicle model. Therefore, we employ a simplified force-constrained vehicle model, for which an analytical solution exists. The core of this vehicle model is a point mass, and suppose a kinematic steering relation to map the steering angle to the corresponding lateral acceleration. This lateral acceleration is converted into a lateral force by multiplying by the vehicle's mass. Similarly, the overall longitudinal force is obtained by dividing the wheel torque by the wheel's rolling radius.

In order to saturate the available acceleration potential and provide an analytical solution of the g-g surface, we constrain the combined force acting on the vehicle using the following equations:
\begin{align}
     F_{x,\mathrm{max}} &= F_{\mathrm{max}} \label{eq:Fx_max_analy}\\
     F_{y,\mathrm{max}} &= \sqrt{F_{\mathrm{max}}^2 - F_{x,\mathrm{drag}}^2} \label{eq:Fy_max_analy}
\end{align}

\begin{figure}[!tb]
\centering
\includegraphics[width=0.95\columnwidth]{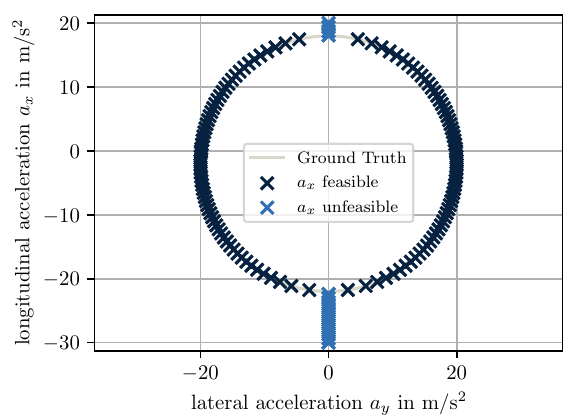}
\caption{Simulation result for a simplified force-constrained vehicle model. Each cross represents one simulation run, while the analytical solution of the g-g diagram is depicted as a solid line. For these simulations, a drag force corresponding to a longitudinal acceleration of $2 \, \mathrm{m/s}^2$ was assumed. The detection method proved to be accurate with very small error margins: the average lateral acceleration ($a_y$) error is below $10^{-5} \, \mathrm{m/s^2}$.}
\label{fig:validation}
\end{figure}

Here, \(F_{\mathrm{max}}\) denotes the total available force envelope, while \(F_{x,\mathrm{drag}}\) emulates a drag force component effectively shifting the circle downwards by \(2 \,\, \si{m/s^2}\).
The constraints \eqref{eq:Fx_max_analy}-\eqref{eq:Fy_max_analy} result in a circular g-g envelope. This analytical solution provides a baseline against which our method can be validated.
As illustrated in Fig.~\ref{fig:validation}, the distribution of simulated data points (crosses) closely follows the analytical boundary (solid line), confirming that our method accurately reproduces the theoretical circular g-g envelope. The tested range for the set of longitudinal accelerations \(\{a_{x,i}\}\) was chosen as 80 uniformly spaced values in the interval
\[
a_{x,i} \in [-30,\,20] \,\,\mathrm{m/s}^2, \quad i = 1, 2, \ldots, 80.
\]
If the velocity-tracking PID controller was not able to settle at the target speed, the corresponding \(a_{x,i}\) was deemed unfeasible, which is labeled as \(a_{x,\text{unfeasible}}\).

\subsection{Extension of a Two-Track Model with Variable Aerodynamics} \label{sec:two_track_model_aero}
We demonstrate how the proposed black box approach scales with added model complexity by extending the two-track model from \cite{sagmeister2024} to include speed- and pitch-dependent aerodynamic forces. The lift coefficient \(c_l\) is adapted to increase with speed, while the aerodynamic balance is modeled to move forward as the vehicle pitches forward. These dependencies are represented as one-dimensional look-up functions in place of constant aerodynamic parameters.
The same quasi-steady simulations described in Section~\ref{sec:method} are run with this extended model. 
\begin{figure}
\centering
\includegraphics[width=0.95\columnwidth]{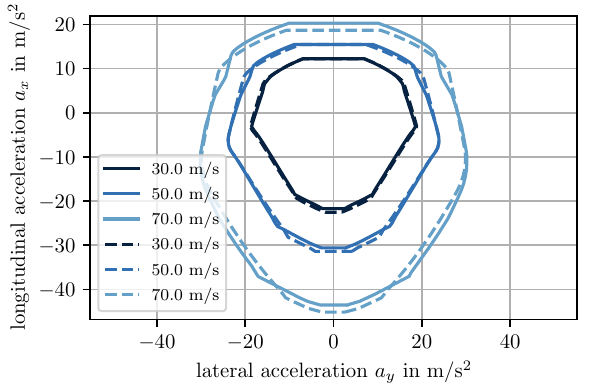}
\caption{Comparison of the results from the extended two-track model with speed- and pitch-dependent aerodynamics (solid lines) versus the baseline model with fixed aerodynamics (dotted lines). at different velocities.}
\label{fig:aero_comparison}
\end{figure}
Figure~\ref{fig:aero_comparison} compares results from the two-track model with constant aerodynamics and the extended version with variable aerodynamics. Although overall the differences remain small, three noteworthy effects emerge:
\begin{itemize}
    \item Longitudinal Potential (Forward Acceleration): 
    At higher speeds, the increased rear downforce of the extended model yields higher longitudinal accelerations.
    \item Longitudinal Potential (Braking): 
    Greater front downforce and therefore reduced rear downforce during braking reduces the negative longitudinal acceleration potential, because the brake balance was not adjusted between the simulation runs.
    \item Combined Maneuvers: 
    When lateral and longitudinal forces overlap, the mechanical balance interacts with the shifting aerodynamic balance, resulting in a mixed picture of improvements or reductions depending on the speed and the pitch angle.
\end{itemize}

This example highlights that our approach can be used to evaluate the impact of additional subsystems, like enhanced aerodynamics, which can change the vehicle performance in both pure and combined acceleration scenarios, without requiring further adaptations in the simulation process. If additional complex subsystems (such as active differentials or electronic stability control) were integrated, the same black box procedure would still apply, avoiding the need to re-implement these subsystems for optimization-based methods. This illustrates how the proposed approach readily accommodates diverse vehicle subsystems and model complexity, while avoiding the model mismatch and convergence issues that typically occur in optimization-based g-g generation methods.

\section{DISCUSSION}
The results confirm that our QSS ramp steer approach avoids unstable driving states, a problem encountered in previous work with QSS analyses \cite{Tremlett.2014}.  Employing a black box simulation interface also allows the integration of proprietary or complex vehicle models without reformulating them for optimization solvers. This also opens the possibility of evaluating neural network-based or other types of vehicle models, besides classical vehicle models. However, since our approach is simulation-based, the validity of the output relies on accurate parameter identification and validation of the vehicle model itself.

An important consideration for efficiently discretizing the longitudinal acceleration dimension, \( a_x \), is the choice of a suitable sampling strategy. A uniform grid in \( a_x \) guarantees a predefined coverage of the domain, however it can lead to a high number of simulations and reduced resolution especially towards the upper and lower end of the diagram. This issue can be observed in Fig.~\ref{fig:validation} as the variation in the maximum lateral acceleration \( a_{y,\mathrm{max}} \) between consecutive \( a_x \) steps tends to increase in these regions, where a more adaptive sampling approach would increase resolution.

\section{CONCLUSION \& OUTLOOK}
This paper presented a novel open-source simulation-based approach to generate the maximum feasible acceleration space for a vehicle, in the form of g-g-g-v diagrams. Our method employs virtual external forces to emulate the effects of longitudinal and vertical accelerations, and performs quasi-steady-state ramp steer maneuvers to determine the maximum feasible lateral accelerations at different vehicle speeds.
The approach enables the use of high-fidelity or black box vehicle models without the need for model differentiability, which is an advantage over existing optimization-based methods.

We validated our method by computing the g-g diagram for a simplified force-constrained vehicle model, and demonstrated its scalability by deriving the g-g-g-v envelope for a high-fidelity two-track race car model with variable aerodynamics. 
The results show that our approach accurately captures the open-loop stable g-g-g-v boundaries, and can be used to quickly evaluate the impact of additional vehicle subsystems on the vehicle's performance.

Future work will assess how the tire-road friction scales the g-g-g-v diagram, and will further improve the efficiency and accuracy of the simulation toolchain, by implementing adaptive sampling strategies for the longitudinal acceleration dimension.

    
\bibliographystyle{IEEEtran}
\bibliography{root}

\end{document}